\documentclass[runningheads]{llncs}

\usepackage{fullpage}
\usepackage{microtype}
\usepackage{soul}
\usepackage{url}
\usepackage{graphicx}
\usepackage[utf8]{inputenc}
\usepackage{vwcol}
\usepackage{hyperref}
\usepackage[inline]{enumitem}
\usepackage[linesnumbered,ruled,vlined]{algorithm2e}
\usepackage{subcaption}
\usepackage{amsmath}
\usepackage{amssymb}
\usepackage{paralist}
\usepackage{appendix}
\usepackage{rotating}
\usepackage{amsmath,amsfonts,bm}

\def\eqref#1{equation~\ref{#1}}
\def\1{\bm{1}}

\newcommand{\test}{\mathcal{D_{\mathrm{test}}}}

\def\eps{{\epsilon}}

\DeclareMathAlphabet{\mathsfit}{\encodingdefault}{\sfdefault}{m}{sl}
\SetMathAlphabet{\mathsfit}{bold}{\encodingdefault}{\sfdefault}{bx}{n}

\newcommand{\R}{\mathbb{R}}

\def\eps{\epsilon}
\def\R{\mathbb{R}}

\def\suchthat{\;:\;}

\newcommand{\abs}[1]{\left|#1\right|}
\newcommand{\norm}[1]{\left\|#1\right\|_{2}}
\newcommand{\infnorm}[1]{\left\|#1\right\|_{\infty}}

\newcommand{\prob}[1]{\operatorname{Pr} \left(#1\right)}
\newcommand{\expec}[1]{{\mathbb E} \left[#1\right]}

\newcommand{\tr}[1]{\operatorname{tr}\left(#1\right)}
\newcommand{\inner}[2]{\left\langle#1,#2\right\rangle}
\newcommand{\eat}[1]{}

\begin{document}
%
%\title{Universal Adversarial Attack using Very Few Test Examples}
\title{Universalization of Any Adversarial Attack using Very Few Test Examples}
%
%\titlerunning{Abbreviated paper title}
% If the paper title is too long for the running head, you can set
% an abbreviated paper title here
%
\author{Sandesh Kamath\inst{1,2} \and
Amit Deshpande\inst{3} \and
K V Subrahmanyam\inst{2} \and 
Vineeth N Balasubramanian\inst{1}}
%
%\authorrunning{Kamath et al.}
% First names are abbreviated in the running head.
% If there are more than two authors, 'et al.' is used.
%
\institute{Indian Institute of Technology, Hyderabad \and Chennai Mathematical Institute, Chennai, India \and
Microsoft Research, Bengaluru, India}
%\email{{ksandeshk,kv}@cmi.ac.in}\\
%\email{amitdesh@microsoft.com}}
%}
%
\maketitle              % typeset the header of the contribution
\begin{abstract}
Deep learning models are known to be vulnerable not only to input-dependent adversarial attacks but also to input-agnostic or universal adversarial attacks. Dezfooli et al.~\cite{Dezfooli17,Dezfooli17anal} construct universal adversarial attack on a given model by looking at a large number of training data points and the geometry of the decision boundary near them. Subsequent work~\cite{Khrulkov18} constructs universal attack by looking only at test examples and intermediate layers of the given model. In this paper, we propose a simple universalization technique to take any input-dependent adversarial attack and construct a universal attack by only looking at very few adversarial test examples. We do not require details of the given model and have negligible computational overhead for universalization. We theoretically justify our universalization technique by a spectral property common to many input-dependent adversarial perturbations, e.g., gradients, Fast Gradient Sign Method (FGSM) and DeepFool. Using matrix concentration inequalities and spectral perturbation bounds, we show that the top singular vector of input-dependent adversarial directions on a small test sample gives an effective and simple universal adversarial attack. For standard models on CIFAR10 and ImageNet, our simple universalization of Gradient, FGSM, and DeepFool perturbations using a test sample of 64 images gives fooling rates comparable to state-of-the-art universal attacks~\cite{Dezfooli17,Khrulkov18} for reasonable norms of perturbation.
\end{abstract}

\begin{figure}
\centering
  \includegraphics[width=0.3\textwidth]{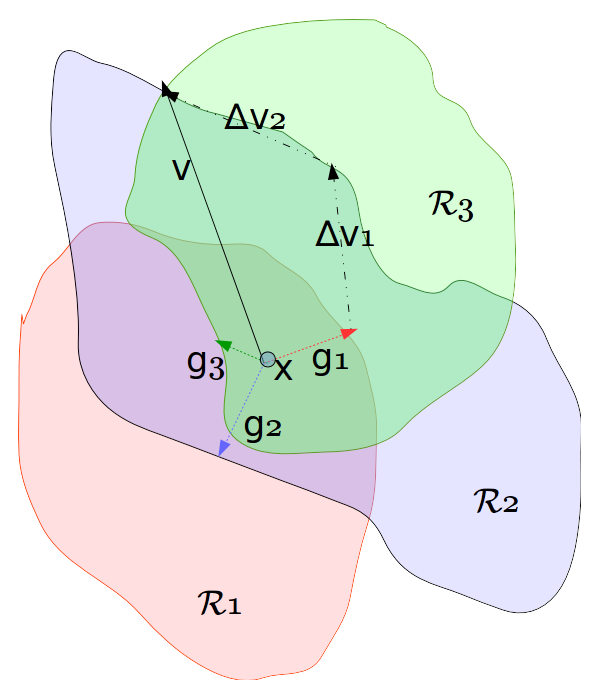}
  \caption{Illustration of the universal adversarial attack problem.}
  %\Description{.}
  \label{fig:universal-fig}
\end{figure}
\section{Introduction} \label{sec:intro}
Neural network models achieve high accuracy on several image classification tasks but are also known to be vulnerable to adversarial attacks. Szegedy et al.~\cite{Szegedy13} showed that tiny pixel-wise changes in images, although imperceptible to the human eye, make highly accurate neural network models grossly misclassify. For a given classifier $f$, an adversarial attack $\mathcal{A}$ perturbs each input $x$ by a carefully chosen small perturbation $\mathcal{A}(x)$ that changes the predicted label as $f(x+\mathcal{A}(x)) \neq f(x)$, for most inputs. Most adversarial attacks are \emph{input-dependent}, i.e., $\mathcal{A}(x)$ depends on $x$. If the underlying model parameters $\theta$ for the classifier $f$ are trained to minimize certain loss function $L(\theta, x, y)$ on data point $x$ with label $y$, then perturbing along the gradient $\nabla_{x} L(\theta, x, y)$ is a natural adversary for maximizing loss, and hopefully, changing the predicted label. If an adversarial attack $\mathcal{A}$ changes each pixel value by at most $\pm \eps$, then its $\ell_{\infty}$-norm is bounded as $\infnorm{\mathcal{A}(x)} \leq \eps$, for all $x$. Szegedy et al.~\cite{Szegedy13} showed that it is possible to find such a perturbation using box-constrained L-BFGS. Goodfellow et al ~\cite{Goodfellow15} proposed the Fast Gradient Sign Method (FGSM) using $\mathcal{A}(x) = \epsilon~ \text{sign}\left(\nabla_{x} L(\theta, x, y)\right)$ as a faster approach to find such an adversarial perturbation. Subsequent work on FGSM includes an iterative variant by Kurakin et al ~\cite{Kurakin17} and another version called Projected Gradient Descent (PGD) by Madry et al ~\cite{Madry18}, both of which constructed adversarial perturbations with bounded $\ell_{\infty}$-norm. On the other hand, DeepFool by Moosavi-Dezfooli et al.~\cite{Dezfooli16} computed a minimal $\ell_{2}$-norm adversarial perturbation iteratively. In each iteration, it used a polyhedron $\mathcal{P}_{t}$ to approximate a region around the current iterate $x^{(t)}$, where the classifier output is the same as $f(x^{(t)})$. The next iterate $x^{(t+1)}$ was the projection of $x^{(t)}$ on to the nearest face of $\mathcal{P}_{t}$. The algorithm was terminated when $f(x^{(t)}) \neq f(x)$, so the perturbation produced by DeepFool on input $x$ is $x^{(t)} - x$. All \emph{input-dependent} adversarial attacks mentioned above can be executed at test time with access only to the given model but not its training data. 

%\begin{figure}[!h]
%\begin{center}
%\includegraphics[width=0.3\linewidth]{images/universal.png}
%\caption{Illustration of the universal adversarial attack problem.}
%\label{fig:universal-fig}
%\end{center}
%\end{figure}

Universal adversarial perturbations are \emph{input-agnostic}, i.e., a given model gets fooled into misclassification by the same perturbation on a large fraction of inputs. Moosavi-Dezfooli et al. \cite{Dezfooli17} constructed a universal adversarial attack by clever, iterative calls to their input-dependent DeepFool attack. They theoretically justified the phenomenon of universal adversarial perturbations using certain geometric assumptions about the decision boundary \cite{Dezfooli17anal}. Given a data distribution $\mathcal{D}$, a universal adversarial perturbation is a vector $v$ of \emph{small} $\ell_{2}$-norm such that $f(x+v) \neq f(x)$, with high probability (called the \emph{fooling rate}), for test input $x$ sampled from the distribution $\mathcal{D}$. For a given bound $\epsilon$ on the $\ell_{2}$-norm of universal adversarial perturbation and a given desired fooling rate, Moosavi-Dezfooli et al. \cite{Dezfooli17} considered a sample $S$ of training data, initialized $v = \bar{0}$, and proceeded iteratively as follows: if the fraction of $x \in S$ for which $f(x+v) \neq f(x)$ is less than the desired fooling rate, then they pick an $x$ such that $f(x+v) = f(x)$, and find a minimal $\ell_{2}$-norm perturbation $\Delta v_{x}$ such that $f(x + v + \Delta v_{x}) \neq f(x + v)$ using DeepFool. Then they update $v$ to $v + \Delta v_{x}$ and scale it down, if required, to have its $\ell_{2}$-norm bounded by $\epsilon$. Figure \ref{fig:universal-fig} gives an illustration of their approach on points $x_{1}, x_{2}, x_{3}$ belonging to distinct classes, shown in three colors. For visualization purposes, the regions containing these points are shown overlapped at points $x_1,x_2,x_3$,  which is the point labeled $x$ in the figure. Let $g_{1}, g_{2}, g_{3}$ be the minimal $\ell_{2}$-norm perturbations such that $f(x_{i} + g_{i}) \neq f(x_{i})$. Moosavi-Dezfooli et al. \cite{Dezfooli17} iteratively identified adversarial perturbations $\Delta v_{1}$ and $\Delta v_{2}$ such that $f(x_{2} + g_{1} + \Delta v_{1}) \neq f(x_{2})$ and $f(x_{3} + g_{1} + \Delta v_{1} + \Delta v_{2}) \neq f(x_{3})$. In Figure \ref{fig:universal-fig}, $v = g_{1} + \Delta v_{1} + \Delta v_{2}$ achieves $f(x_{i} + v) \neq f(x_{i})$, for $i=1, 2, 3$, simultaneously. Moosavi-Dezfooli et al. \cite{Dezfooli17} showed that the universal adversarial perturbation constructed as above from a large sample of training data gave a good fooling rate even on test data. Note that the above construction of universal adversarial attack requires access to training data and several iterations of the DeepFool attack.

The above discussion raises some natural, important questions: \begin{enumerate*}[label=(\alph*)] \item Is there a simpler construction to \emph{universalize} any given input-dependent adversarial attack? \item Can a universal attack be constructed efficiently using access to the model and very few test inputs, with no access to the training data at all (or the test data in entirety)? \end{enumerate*} We answer both of these questions affirmatively.
%We answer these affirmatively using the following cues: \begin{inparaenum}[(a)] \item Can we exploit any interesting properties common to input-dependent adversarial attacks for a simpler construction of universal adversarial perturbation?  \item Can we find a universal adversarial perturbation for a given model using only a small fraction of test inputs so that this perturbation fools a large fraction of test inputs? \end{inparaenum}

Our key results are summarized as follows:

%\noindent \textbf{Summary of our results}
\noindent $\bullet$ Our first observation is that many known input-dependent adversarial attack directions have only a small number of dominant principal components on the entire data. We firstly show this for attacks based on the gradient of the loss function, the FGSM attack, and DeepFool, on different architectures and datasets. %However, we report the findings only on ImageNet.

\noindent $\bullet$ Consider a matrix whose each row corresponds to input-dependent adversarial direction for a test data point. Our second observation is that a small perturbation along the top principal component of this matrix is an effective universal adversarial attack. This simple approach using Singular Value Decomposition (SVD), our SVD-Universal algorithm combined with Gradient, FGSM and DeepFool directions gives us SVD-Gradient, SVD-FGSM and SVD-DeepFool universal adversarial attacks, respectively.

\noindent $\bullet$ Our third observation is that the top principal component can be well-approximated from a very small sample of the test data (following \cite{Khrulkov18}), and SVD-Universal approximated from even a small sample gives a fooling rate comparable to Moosavi-Dezfooli et al.~\cite{Dezfooli17}. Importantly, this approach can be used with any attack, as we show with three different methods in this work.
%\eat{SVD-Gradient and SVD-DeepFool obtained from a $0.2\%$-sample of ImageNet validation data when scaled up to $\ell_2$-norm 25 (this is 0.06 of the average $\ell_2$ norm of the dataset, 450), give small pixel-wise perturbations, and fool 26\% of the validation data on VGG16.  For CIFAR-10 trained on ResNet18 we get a fooling rate of 35\% using SVD-Gradient perturbations with $\ell_2$ norm 8, obtained from a 1\% sample of the test data (the perturbation is \textcolor{red}{00} of the average norm of the dataset).}

\noindent $\bullet$ We give a theoretical justification of this phenomenon using matrix concentration inequalities and spectral perturbation bounds. This observation holds across multiple input-dependent adversarial attack directions given by Gradient, FGSM and DeepFool.
% Khrulkov and Oseledets ~\cite{Khrulkov18} propose a different method to construct a universal adversarial perturbation using a completely trained model and test data. Their perturbation is based on computing the $(p,q)$-singular vectors of the Jacobian matrices of hidden layers of the network.

\section{Related Work}
%In Moosavi-Dezfooli et al.~\cite{Dezfooli17}, the authors fix a budget $\epsilon$, a bound on the allowed $\ell_2$ norm of the attack\footnote{the authors consider other $\ell_p$ norms as well. We restrict ourselves to the $\ell_2$-norm.} and also the desired fooling rate. They pick a sample $S$ of the training data.  Starting with the zero perturbation $v$ the algorithm proceeds in stages. If the fooling rate of the current perturbation vector $v$ is less than the desired fooling rate on $S$, they run over $x \in S$, for which $f(x + v) = f(x)$.  For each such $x$ they find the smallest perturbation $r_x$ sending $x$ to the boundary (using DeepFool) and $v$ is updated to $\Pi(v + r_x)$ where $\Pi$ is the projection onto the ball of radius $\epsilon$.
The previous works closest to ours are the universal adversarial attacks by Moosavi-Dezfooli et al. \cite{Dezfooli17} and Khrulkov and Oseledets \cite{Khrulkov18}. Our approach to construct a universal adversarial attack is to take an input-dependent adversarial attack on a given model, and then find a single direction via SVD that is simultaneously well-aligned with the different input-dependent attack directions for most test data points. This is different from the approach of Moosavi-Dezfooli et al. \cite{Dezfooli17} explained in Figure \ref{fig:universal-fig}. When a training data point is not fooled by a smaller perturbation in previous iterations, Moosavi-Dezfooli et al. \cite{Dezfooli17} apply DeepFool to such already-perturbed but unfooled data points. In contrast, our universalization uses input-dependent attack directions only on a small sample of data points, and even the simple universalization of gradient directions (instead of DeepFool) already gives a comparable fooling rate to the universal attack of Moosavi-Dezfooli et al. \cite{Dezfooli17} in our experiments.

Khrulkov and Oseledets~\cite{Khrulkov18} propose a state-of-the-art universal adversarial attack that requires expensive computation and access to the hidden layers of the given neural network model. They consider the function $f_{i}(x)$ computed by the $i$-th hidden layer on input $x$, and its Jacobian $J_{i}(x) = \partial f_{i}/\partial x\big|_{x}$. Using $\|f_{i}(x+v) - f_{i}(x)\|_{q} \approx \|J_{i}(x)v\|_{q}$, they solve a $(p, q)$-SVD problem to maximize $\sum_{x \in \mathcal{X}} \left\|J_{i}(x)v\right\|_{q}^{q}$ subject to $\left\|\right\|_{p} = 1$ over the entire data $\mathcal{X}$. They optimize for the choice of layer $i$ and $(p, q)$ in the $(p, q)$-SVD for the $i$-th hidden layer. They \emph{hypothesize} that this objective can be empirically approximated by a sample $S$ of size $m$ from test data (see Eqn.(8) in \cite{Khrulkov18}). With extensive experiments on ILSVRC 2012 validation data for VGG-16, VGG-19 and ResNet50 models, they empirically find the best layer $i$ to attack and the empirically best choice of $(p, q)$ (e.g., $q=10$ for $p=\infty$). We do not solve the general $(p, q)$-SVD, which is known to be NP-hard for most choices of $(p, q)$ \cite{Bhaskara11,Bhattiprolu19}. Our SVD-Universal algorithm uses the regular SVD ($p=q=2$), which can be solved provably and efficiently. Our method does not require access to hidden layers, and it universalizes several known input-dependent adversarial perturbations. We \emph{prove} that our objective can be well-approximated from only a small sample of test data (Theorem \ref{thm:sample}), following Khrulkov and Oseledets \cite{Khrulkov18}, who however only \emph{hypothesize} this for their objective (see Eqn.(8) in \cite{Khrulkov18}).

Recent work has also considered model-agnostic and data-agnostic adversarial perturbations. Tramer et al.~\cite{Tramer17} study model-agnostic perturbations in the direction of the difference between the intra-class means, and come up with adversarial attacks that transfer across different models. Mopuri et al.~\cite{Mopuri17} propose a data-agnostic adversarial attack that depends only on the model architecture. Given a trained neural network with $k$ hidden (convolution) layers, they start with a random image $v$ and minimize $\prod_{i=1}^{k} \ell_{i}(v)$ subject to $\|v\|_{\infty} \leq \epsilon$, where $\ell_{i}(v)$ is the mean activation of the $i$-th hidden layer for input $v$. The authors show that the optimal perturbation $v$ for this objective exhibits a data-agnostic adversarial attack. In contrast to these methods, we present a simple yet effective method based on the principal component of a few attack directions, as described further below.

\section{SVD-Universal: A Simple Method to Universalize an Adversarial Attack} \label{sec:univ}
We begin by defining the notation and the evaluation metric \emph{fooling rate} formally. Let ${\cal D}$ denote the data distribution on image-label pairs $(x, y)$, with images as a $d$-dimensional vectors in some ${\mathcal X} \subseteq \R^{d}$ and labels in $[k] = \{1, 2, \dotsc, k\}$ for $k$-class classification, e.g., CIFAR-10 data has images with $32 \times 32$ pixels that are essentially $1024$-dimensional vectors of pixel values, each in $[0, 1]$, along with their respective labels for $10$-class classification. Let $(X,Y)$ be a random data point from ${\cal D}$ and let $f:{\cal X} \rightarrow [k]$ be a $k$-class classifier. We use $\theta$ to denote the model parameters for classifier $f$, and let $L(\theta, x, y)$ denote the loss function it minimizes on the training data. The accuracy of classifier $f$ is given by $\text{Pr}_{(X, Y)} \left(f(X) = Y\right)$.
%The error rate of a classifier is  $\text{Pr}_{(X, Y) \in {\cal D}} [f(X) \not = Y]$.\eat{We regard a deterministic adversary to be a function ${\cal A}: {\mathcal X} \rightarrow {\mathcal X}$. The error rate of ${\cal A}$ on the classifier $f$ is defined to be 
%\[ \text{Pr}_{(X, Y) \in {\cal D}}[ f (X + {\cal A}(X)) \neq Y].\]}
% An adversary 
%${\cal A}$ is a function ${\cal X} \rightarrow {\mathbb R}^d$. When ${\cal A}$ is a distribution over functions we get a randomized adversary. \eat{The definition of fooling rate remains the same as above except that we include the randomness in ${\cal A}$ when calculating the above probability.}The norm of the perturbation applied to $X$ is the norm of ${\cal A}(X)$ (we only consider $\ell_2$ norm in this paper).
%
%In Moosavi-Dezfooli et al.~\cite{Dezfooli17,Dezfooli17anal} and Khrulkov and Oseledets~\cite{Khrulkov18}, the authors consider the fooling rate of an adversary. 
A classifier $f$ is said to be fooled on input $x$ by adversarial perturbation ${\cal A}(x)$ if $f(x + {\cal A}(x)) \neq f(x)$. The fooling rate of the adversary ${\cal A}$ is defined as $\text{Pr}_{(X,Y)} \left(f(X + {\cal A}(X)) \neq f(X)\right)$.
%\[ \text{Pr}_{(X,Y)}[f(X + {\cal A}(X)) \neq f(X)].\]
%The error rate of ${\cal A}$ on the classifier $f$ is defined to be 
%$\text{Pr}_{(X, Y) \in {\cal D}}[ f (X + {\cal A}(X)) \neq Y]$. It is easy to see that 
%\begin{equation*}
%\begin{split}
%& \text{Pr}_{(X,Y)}[ f (X + {\cal A}(X)) \neq f(X)] \\
%& \geq  \text{Pr}_{(X, Y)}[ f (X + {\cal A}(X)) \neq Y] - (1 - \delta). \\
%\end{split}
%\end{equation*}
%So, if the natural accuracy of the classifier $f$ is high,
%the fooling rate is close to the error rate.
%The error rate of the adversary with zero perturbation is the error rate of the trained network, whereas the fooling rate of the adversary with zero perturbation is necessarily zero. However, small fooling rate does not necessarily imply small error rate, especially when the natural accuracy is not close to 100\%. Note that existing models such as VGG16, VGG19, ResNet50 do not achieve natural accuracy greater than 0.8 on the ImageNet dataset.

\begin{algorithm}[h]
\SetAlgoLined
\KwData{A neural network $N$, an input-dependent adversarial attack $\mathcal{A}$, and $n$ test samples.}
\KwResult{A universal attack direction for neural network $N$}
 
For test samples $x_{1}, x_{2}, \dotsc, x_{n}$, obtain input-dependent perturbation vectors $a_{1} = \mathcal{A}(x_{1}), a_{2} = \mathcal{A}(x_{2}), \dotsc, a_n = \mathcal{A}(x_{n})$ for the neural network $N$. 

Normalize $a_i$'s to get the attack directions or unit vectors $u_{i} = a_{i}/||a_{i}||_{2}$, for $i=1$ to $n$.

Form a matrix $M$ whose rows are $u_{1}, u_{2}, \dotsc, u_{n}$.

Compute Singular Value Decomposition (SVD) of $M$ as $M = USV^T$, with $V = [v_{1}|v_{2}|\dotsc|v_{n}]$.

Return the top right singular vector $v_{1}$ as the universal attack vector. 
\caption{SVD-Universal Algorithm}
\label{alg:svd-univ}
\end{algorithm}

Our approach to construct a universal adversarial attack is to take an input-dependent adversarial attack on a given model, and then find a single direction via SVD that is simultaneously well-aligned with the different input-dependent attack directions for most test data points. We apply this approach to a very small sample (less than $0.2\%$) of test data, and use the top singular vector as a universal adversarial direction. Our algorithm, SVD-Universal, is presented in Algorithm \ref{alg:svd-univ}. We prove that if the input-dependent attack directions satisfy a certain spectral property, then our approach can provably result in a good fooling rate (Theorem \ref{thm:eigen}), and we prove that a small sample size suffices, independent of the data dimensionality (Theorem \ref{thm:sample}).

Our SVD-Universal algorithm is flexible enough to universalize many popular input-dependent adversarial attacks. We apply it in three different ways to construct input-dependent perturbations: \begin{enumerate*}[label=(\alph*)] \item Gradient attack that perturbs an input $x$ in the direction $\nabla_{x} L(\theta, x, y)$, \item FGSM attack \cite{Goodfellow15} that perturbs $x$ in the direction $\text{sign}\left(\nabla_{x} L(\theta, x, y)\right)$, \item DeepFool attack \cite{Dezfooli16} which is an iterative algorithm explained in Section \ref{sec:intro}. \end{enumerate*} For the above three input-dependent attacks, we call the universal adversarial attack produced by SVD-Universal as SVD-Gradient, SVD-FGSM, and SVD-DeepFool, respectively.

%Before presenting the fooling rate achieved by the algorithm, we give details of the augmentation applied to the datasets for training the network. Next, we show that the attack on the network gets better as we train it with larger range of rotation augmentations. We follow this up with a comparison with Moosavi-Dezfooli et al. universal attack.
%We do not optimize our attack to be the best universal attack. Rather we observe a simple spectral property, and exploit it to give a generic recipe that takes input-dependent attack directions on a tiny sample of the test data and creates a universal attack that can fool the model on majority of the test data.

%\subsection{SVD-Universal Algorithm}
%We first give the details of our universalization technique which we refer to as SVD-Universal. We refer to the universal vector output by the technique, using a suffix to SVD-* with the adversarial attack direction used for the computaion e.g. if Gradients are used the universal vector is referred to as SVD-Gradient. 
%In Algorithm \ref{alg:svd-univ} we give the pseudocode to construct a universal attack for a given model. The output of the algorithm is a universal attack vector, and this depends upon the input dependent attack used. 

As shown in Algorithm \ref{alg:svd-univ}, SVD-Universal samples a set of $n$ images from the test (or validation) set. We use the terms batch size and sample size interchangeably. For each of the sampled points, we compute an input-dependent attack direction. We stack these attack directions as rows of a matrix. The top right singular vector of this matrix of attack directions is the universal adversarial direction that SVD-Universal outputs.  

\section{Theoretical Analysis of SVD-Universal}
In this section, we provide a theoretical justification for the existence of universal adversarial perturbations.  Let $(X, Y)$ denote a random sample from $\mathcal{D}$. Let $f: \R^{d} \rightarrow [k]$ be a given classifier, and for any $x \in \R^{d}$, let $\mathcal{A}(x)$ be the adversarial perturbation given by a fixed attack ${\cal A}$, say \emph{FGSM}, \emph{DeepFool}. 

Define $A = \{x \suchthat f(x + \mathcal{A}(x)) \neq f(x)\}$. For any $x \in A$, assume that $x + \mathcal{A}(x)$ lies on the decision boundary, and let the hyperplane $H_{x} = \{x + z \in \R^{d} \suchthat \inner{z}{\mathcal{A}(x)} = \norm{\mathcal{A}(x)}^{2}\}$ be a local, linear approximation to the decision boundary at $x + \mathcal{A}(x)$. This holds for adversarial attacks such as DeepFool by Moosavi-Dezfooli et al.~\cite{Dezfooli16} which try to find an adversarial perturbation $\mathcal{A}(x)$ such that $x + \mathcal{A}(x)$ is the nearest point to $x$ on the decision boundary. Now consider the halfspace $S_{x} = \{x + z \in \R^{d} \suchthat \inner{z}{\mathcal{A}(x)} \geq \norm{\mathcal{A}(x)}^{2}\}$. Note that $x \notin S_{x}$ and $x + \mathcal{A}(x) \in S_{x}$. \emph{For simplicity of analysis, we assume that $f(x + z) \neq f(x)$, for all $x \in A$ and $x + z \in S_{x}$}. This is a reasonable assumption in a small neighborhood of $x$. In fact, this hypothesis is implied by the positive curvature of the decision boundary assumed in the analysis of Moosavi-Dezfooli et al.~\cite{Dezfooli17anal}. Moosavi-Dezfooli et al.~\cite{Dezfooli17anal} empirically verify the validity of this hypothesis.  In other words, we assume that if an adversarial perturbation $\mathcal{A}(x)$ fools the model at $x$, then any perturbation $z$ having a sufficient projection along $\mathcal{A}(x)$, also fools the model at $x$. This is a reasonable assumption in a small neighborhood of $x$.

%\eat{Two important questions are: \begin{inparaenum}[(a)] \item Why does there exist a \emph{universal} vector or direction that works as an adversarial perturbation for many or most data points simultaneously? \item Can we compute such a \emph{universal} adversarial perturbation by looking only at a small subsample of the data?
%\end{inparaenum}} \eat{Given any method that obtains adversarial perturbations for individual data points, if the matrix of these adversarial perturbation directions taken over all data points satisfies a certain property, then we show that the top singular vectors of this matrix are good candidates for \emph{universal} adversarial perturbations that make many points to be misclassified simultaneously.}

\begin{theorem} \label{thm:eigen}
Given any joint data distribution $\mathcal{D}$ on features or inputs in $\R^{d}$ and true labels in $[k]$, let $(X, Y)$ denote a random sample from $\mathcal{D}$. For any $x \in \R^{d}$, let $\mathcal{A}(x)$ be its adversarial perturbation by a fixed input-dependent adversarial attack $\mathcal{A}$. Let
\[
M = \expec{\frac{\mathcal{A}(X)}{\norm{\mathcal{A}(X)}} \frac{\mathcal{A}(X)^{T}}{\norm{\mathcal{A}(X)}}} \in \R^{d \times d},
\]
and $0 \leq \lambda \leq 1$ be the top eigenvalue of $M$ and $v \in \R^{d}$ be the normalized unit eigenvector. Then, for any $0 < \delta < \sqrt{\lambda}$, under the assumption that $f(x + z) \neq f(x)$, for all $x \in A$ and $x + z \in S_{x}$, we have\\
%\begin{center}
$$\prob{f(X + u) \neq f(X)} \geq \prob{f(X + \mathcal{A}(X)) \neq f(X)} - \frac{1 - \lambda}{1 - \delta^{2}},$$
%\end{center}
%\begin{equation*}
%\begin{split}
%\prob{f(X + u) \neq f(X)} & \geq \\
%& \hspace{-0.3in} \prob{f(X + \mathcal{A}(X)) \neq f(X)} - \frac{1 - \lambda}{1 - \delta^{2}},
%\end{split}
%\end{equation*}
where $u = \pm (\epsilon/\delta) v$, where $\epsilon = \underset{x}{\max} \norm{\mathcal{A}(x)}$. 
\end{theorem}
\begin{proof}
Let $\mu(x)$ denote the induced probability density on features or inputs by the distribution $\mathcal{D}$. Define $A = \{x \suchthat f(x + \mathcal{A}(x)) \neq f(x)\}$ and $G = \{x \suchthat \abs{\inner{\mathcal{A}(x)}{v}} \geq \delta~ \norm{\mathcal{A}(x)}\}$. Since $\lambda$ is the top eigenvalue of $M$ with $v$ as its corresponding (unit) eigenvector,
\begin{align*}
\lambda & = \expec{\inner{\frac{\mathcal{A}(X)}{\norm{\mathcal{A}(X)}}}{v}^{2}} \\
%& = \int_{x \in G} \inner{\frac{\mathcal{A}(x)}{\norm{\mathcal{A}(x)}}}{v}^{2} \mu(x) dx + \int_{x \notin G} \inner{\frac{\mathcal{A}(x)}{\norm{\mathcal{A}(x)}}}{v}^{2} \mu(x) dx \\
& = \int_{x \in G} \inner{\frac{\mathcal{A}(x)}{\norm{\mathcal{A}(x)}}}{v}^{2} \mu(x) dx + \int_{x \notin G} \inner{\frac{\mathcal{A}(x)}{\norm{\mathcal{A}(x)}}}{v}^{2} \mu(x) dx \\
%& + \int_{x \notin G} \inner{\frac{\mathcal{A}(x)}{\norm{\mathcal{A}(x)}}}{v}^{2} \mu(x) dx \\
& \leq \int_{x \in G} \mu(x) dx + \delta^{2}~ \int_{x \notin G} \mu(x) dx \quad \text{because $\norm{v} = 1$} \\
& = \prob{G} + \delta^{2}~ (1 - \prob{G}) \\
& = (1 - \delta^{2})~ \prob{G} + \delta^{2}. 
\end{align*}
Thus, $\prob{G} \geq (\lambda - \delta^{2})/(1 - \delta^{2})$, and equivalently, $\prob{G^{c}} = 1 - \prob{G} \leq (1 - \lambda)/(1 - \delta^{2})$. Now for any $x \in G$, we have $\abs{\inner{\mathcal{A}(x)}{v}} \geq \delta~ \norm{\mathcal{A}(x)}$. Letting $\epsilon = \max_{x} \norm{\mathcal{A}(x)}$, we get $\abs{\inner{\mathcal{A}(x)}{(\epsilon/\delta) v}} \geq \norm{\mathcal{A}(x)}^{2}$. Thus, $x \pm (\epsilon/\delta) v \in S_{x}$, where $S_{x} = \{x + z \in \R^{d} \suchthat \inner{z}{\mathcal{A}(x)} \geq \norm{\mathcal{A}(x)}^{2}\}$, and therefore, by our assumption stated before Theorem \ref{thm:eigen}, we have $f(x) \neq f(x + u)$, where $u = \pm (\epsilon/\delta) v$. Putting all of this together, $\prob{f(X + u) \neq f(X)} \geq \prob{G \cap A} \geq \prob{A} - \prob{A \cap G^{c}} \geq \prob{A} - \prob{G^{c}}$, and therefore,\\
$\prob{f(X + u) \neq f(X)} \geq \prob{f(X + \mathcal{A}(X)) \neq f(X)} - \frac{1 - \lambda}{1 - \delta^{2}}$.
%\begin{equation*}
%\begin{split}
%\prob{f(X + u) \neq f(X)} & \geq\\
%& \hspace{-0.2in} \prob{f(X + \mathcal{A}(X)) \neq f(X)} - \frac{1 - \lambda}{1 - \delta^{2}}.
%\end{split}
%\end{equation*}
\end{proof}

\begin{figure}[!h]
\begin{center}
\includegraphics[width=0.49\linewidth]{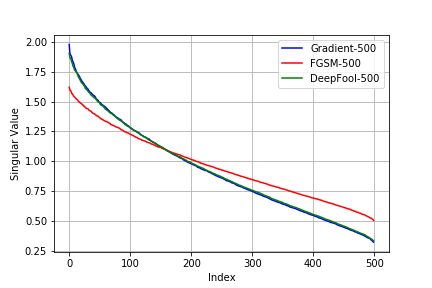}
\includegraphics[width=0.49\linewidth]{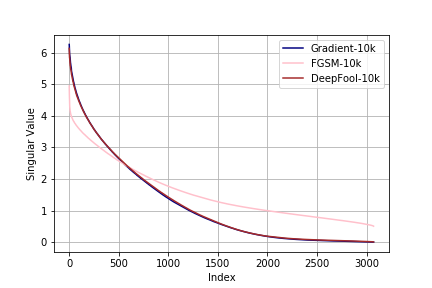} 
\caption{On CIFAR-10, ResNet18, Singular values of attack directions over a sample of (top) 500 and (bottom) 10,000 test points.}
\label{fig:cifar10-cnn-sing-val}
\end{center}
\end{figure}

Theorem \ref{thm:eigen} shows that any norm-bounded, input-dependent, adversarial attack $\mathcal{A}$ can be converted into a universal attack $u$ of comparable norm, without losing much in the fooling, if the top eigenvalue $\lambda$ of $M$ is close to $1$. This universal attack direction lies in the one-dimensional span of the top eigenvector $v$ of $M$. The proof of Theorem \ref{thm:eigen} can be easily generalized to the top SVD subspace of $M$ and where the top few eigenvalues of $M$ dominate its spectrum (note that $\tr{M} = 1$).

%\eat{\textcolor{red}{Observe that $\tr{M} = 1$ since this is true of each term inside the expection. Note that because we are taking an expectation, the eigen values of $M$ are normalized and so at most one.}  
%Theorem \ref{thm:eigen} implies that as the top eigenvalue $\lambda$ dominates the spectrum, its eigenvector $v$ get more aligned with the adversarial perturbations directions $a_{x}/\norm{a_{x}}$ for most points $x \in X$. 
%This means that $v$ is a potential candidate for a \emph{universal} adversarial perturbation. More generally, Theorem $\ref{thm:eigen}$ works for any of the top eigenvalues of $M$ and their corresponding eigenvectors, and thus, gives a subspace spanned by multiple orthogonal directions all of which are potential candidates for being \emph{universal} adversarial perturbations that fool a given classifier on many input examples.}

\noindent \textbf{Singular value drop.}
We empirically verify our hypothesis about top eigenvalue (or the top few eigenvalues) dominating the spectrum of $M$ in Theorem~\ref{thm:eigen}. Let $X_{1}, X_{2}, \dotsc, X_{m}$ be $m$ i.i.d. samples of $X$ drawn from the distribution $\mathcal{D}$ and consider the unnormalized, empirical analog of $M$ as follows:
\[
\sum_{i=1}^{m} \frac{\mathcal{A}(X_{i})}{\norm{\mathcal{A}(X_{i})}} \frac{\mathcal{A}(X_{i})^{T}}{\norm{\mathcal{A}(X_{i})}}.
\]
Figure (\ref{fig:cifar10-cnn-sing-val}) shows how the singular values drop for the three input dependent attacks, {\it Gradient}, {\it FGSM}, and {\it DeepFool} on CIFAR-10 trained on ResNet18 on batch sizes 500 and 10,000. These plots indicate that the drop in singular values is a shared phenomenon across different input-dependent attacks, and the trend is similar even when we look at a small number of input samples.

Our second contribution is finding a good approximation to the \emph{universal} adversarial perturbation given by the top eigenvector $v$ of $M$, using only a small sample $X_{1}, X_{2}, \dotsc, X_{m}$ from $\mathcal{D}$. Theorem \ref{thm:sample} shows that we can efficiently pick such a small sample whose size is independent of $\mathcal{D}$, depends linearly on the \emph{intrinsic dimension} of $M$, and logarithmically on the feature dimension $d$.
\begin{theorem} \label{thm:sample}
Given any joint data distribution $\mathcal{D}$ on features in $\R^{d}$ and true labels in $[k]$, let $(X, Y)$ denote a random sample from $\mathcal{D}$. For any $x \in \R^{d}$, let $\mathcal{A}(x)$ denote the adversarial perturbation of $x$ according a fixed input-dependent adversarial attack $\mathcal{A}$. Let
\[
M = \expec{\frac{\mathcal{A}(X)}{\norm{\mathcal{A}(X)}} \frac{\mathcal{A}(X)^{T}}{\norm{\mathcal{A}(X)}}} \in \R^{d \times d}.
\]
Let $0 \leq \lambda = \norm{M} \leq 1$ denote the top eigenvalue of $M$ and let $v$ denote its corresponding eigenvector (normalized to have unit $\ell_{2}$ norm). Let $r = \tr{M}/\norm{M}$ be the \emph{intrinsic dimension} of $M$. Let $X_{1}, X_{2}, \dotsc, X_{m}$ be $m$ i.i.d. samples of $X$ drawn from the distribution $\mathcal{D}$, and let $\tilde{\lambda} = \norm{\tilde{M}}$ be the top eigenvalue of the matrix $\tilde{M}$,
\[
\tilde{M} = \frac{1}{m} \sum_{i=1}^{m} \frac{\mathcal{A}(X_{i})}{\norm{\mathcal{A}(X_{i})}} \frac{\mathcal{A}(X_{i})^{T}}{\norm{\mathcal{A}(X_{i})}},
\]
and $\tilde{v}$ be the top eigenvector of $\tilde{M}$.

Also suppose that there is a gap of at least $\gamma \lambda$ between the top eigenvalue $\lambda$ and the second eigenvalue of $M$. Then for any $0 \leq \epsilon < \gamma$ and $m = O(\epsilon^{-2} r \log d)$, we get $\norm{v - \tilde{v}} \leq \epsilon/\gamma$, with a constant probability. This probability can be boosted to $1-\delta$ by having an additional $\log (1/\delta)$ in the $O(\cdot)$.
\end{theorem}
\begin{proof}
Take $m = O(\epsilon^{-2} r \log d)$. By the covariance estimation bound (see Vershynin~\cite[Theorem 5.6.1] {vershynin2018}) and Markov's inequality, we get that $\norm{M - \tilde{M}} \leq \epsilon \lambda$, with a constant probability.  Applying Weyl's theorem on eigenvalue perturbation \cite[Theorem 4.5.3]{vershynin2018}, we get $\abs{\lambda - \tilde{\lambda}} \leq \epsilon \lambda$. Moreover, if there is gap of at least $\gamma \lambda$ between the first and the second eigenvalue of $M$ with $\gamma > \epsilon$, we can use the Davis-Kahan theorem \cite[Theorem 4.5.5]{vershynin2018}  to bound the difference between the \eat{corresponding} eigenvectors as $\norm{v - \tilde{v}} \leq \epsilon/\gamma$, with a constant probability. Please see Appendix, and the book by \cite{vershynin2018} cited therein, for more details about the covariance estimation bound, Weyl's theorem, and Davis-Kahan theorem.
\end{proof}

The description of the results cited in the proof above are provided in the Appendix for clarity of reading. The theoretical bounds are weaker than our empirical observations on the number of test samples needed for the attack. We wish to highlight again that the bound in Theorem \ref{thm:sample} is independent of the support of underlying data distribution $\mathcal{D}$ and depends logarithmically on the feature dimension $d$. There is more room to tighten our analysis using more properties of the data distribution and the spectral properties of $M$.

%\eat{The theoretical bounds are weaker than our empirical observations on the number of test samples needed for the attack. We wish to highlight again that the bound in Theorem \ref{thm:sample} is independent of the support of underlying data distribution $\mathcal{D}$ and depends logarithmically on the feature dimension $d$. }\eat{There is more room to tighten our analysis using more properties of the data distribution and the spectral properties of $M$.}

\section{Experiments and Results}
%\subsection{Experimental Setup}%Data sets and model architectures used}
%\label{app:arch}
\noindent \textbf{Datasets.} CIFAR-10 dataset consists of $60,000$ images of $32 \times 32$ size, divided into $10$ classes: $40,000$ used for training, $10,000$ for validation and $10,000$ for testing. ImageNet refers to the ILSRVC 2012 dataset \cite{Russakovsky2015} which consists of images of $224 \times 224 $ size, divided into $1000$ classes. All experiments performed on neural network-based models were done using the validation set of ImageNet and test set of CIFAR-10 datasets. 

\vspace{6pt}
\noindent \textbf{Model Architectures.} For the ImageNet based experiments, we use pre-trained networks of VGG16, VGG19 and ResNet50 architectures\footnote{https://pytorch.org/docs/stable/torchvision/models.html}.
For the CIFAR-10 experiments, we use the ResNet18 architecture as in He et al.~\cite{He16}. All of these are popularly used models.
%\subsection{Comparison of Fooling Rates of SVD-Universal with Baselines}%Moosavi-Dezfooli et al.\protect \cite{Dezfooli17}, Khrulkov and Oseledets \protect \cite{Khrulkov18}}
%\subsection{Network setup before applying the algorithm}
%\noindent \textbf{Model Architectures.} 
%We evaluate the fooling rate of SVD-Gradient, SVD-FGSM and SVD-DeepFool on the validation set of the ImageNet dataset on VGG16, VGG19 and ResNet50.
We used off the shelf code available for these architectures. In these architectures pixel intensities of images are scaled down and images are normalized before they are deployed for use in classifiers. Our (unit) attack vectors are constructed using batch size of 64 (0.13\%). We found that SVD-Universal attacks obtained using batch size of 64 perform as well as SVD-Universal attacks obtained using larger batch sizes of 128 and 1024 (see below). We compare the results of our attacks with M-DFFF, which denotes the universal perturbation vecror obtained using the method of Moosavi-Dezfooli et al~\cite{Dezfooli17}. For fair comparison, the same 64 samples used to construct our SVD-Universal vectors were used to obtain the universal perturbation vector, \textcolor{blue}{M-DFFF}, of Moosavi-Dezfooli et al~\cite{Dezfooli17}. This vector was scaled down to get a unit vector $w$ in $\ell_2$ norm. (Higher is better for all the presented results on error rate or fooling rate.)

\begin{figure}[!h]
\begin{center}
\includegraphics[width=0.48\linewidth]{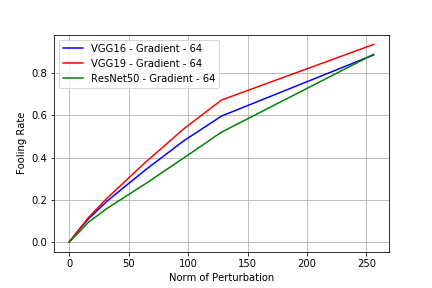}
\includegraphics[width=0.48\linewidth]{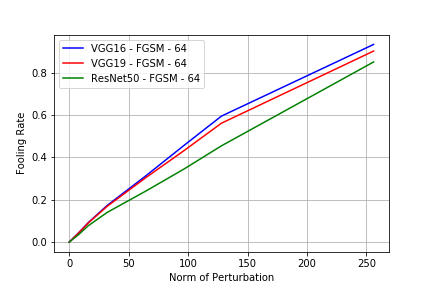}
\includegraphics[width=0.48\linewidth]{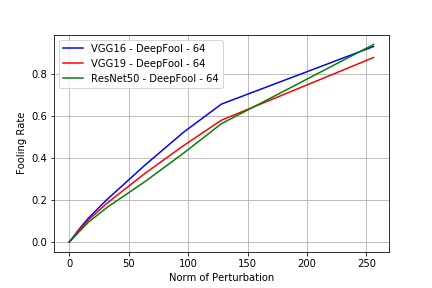}
\includegraphics[width=0.48\linewidth]{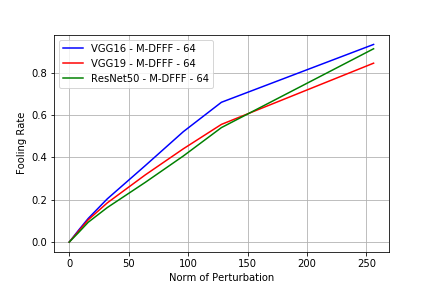}
\end{center}
\caption{On ImageNet validation, VGG16 vs VGG19 vs ResNet50: fooling rate vs. norm of perturbation. Attacks constructed using 64 samples. (top left) {\it SVD-Gradient} (top right)  {\it SVD-FGSM } 
(bottom left) {\it SVD-DeepFool} and (bottom right) M-DFFF universal.}
\label{fig:cnn-fool-imagenet-newdef}
\end{figure}

\begin{figure}[!h]
%\centering
\begin{center}
%\begin{tabular}{ccc}
% \textbf{VGG16} & \textbf{VGG19} & \textbf{ResNet50} \\
\includegraphics[width=0.48\linewidth]{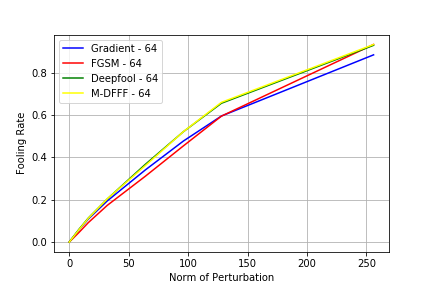} %& 
\includegraphics[width=0.48\linewidth]{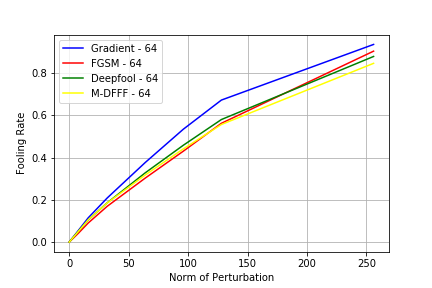} %&
\includegraphics[width=0.48\linewidth]{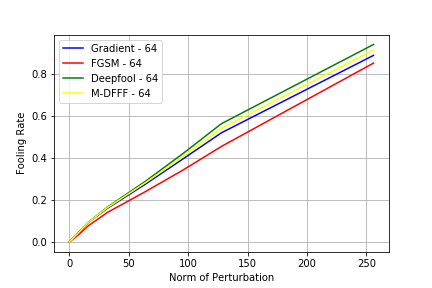}%\\
%\end{tabular}
\end{center}
\caption{On ImageNet validation, \textit{(top left):} VGG16: fooling rate, \textit{(top right)} VGG19: fooling rate, \textit{(bottom)} ResNet50: fooling rate, vs. norm of perturbation along top singular vector of attack directions on 64 samples.}
\label{fig:cnn-fool-vgg16vsvgg19vsres}
\end{figure}

%\subsection{SVD-Universal on ImageNet}\label{app:sec:svd-univ}
\vspace{6pt}
\noindent \textbf{SVD-Universal on ImageNet.}
In Figure \ref{fig:cnn-fool-imagenet-newdef}, each plot has the fooling rate on VGG16, VGG19 and ResNet50 attacked by the same universal method. This shows the effectiveness of the same attack method on different networks. In Figure \ref{fig:cnn-fool-vgg16vsvgg19vsres}, to compare SVD-Universal and M-DFFF, we plot the fooling rate of both together for each network. Importantly, Figure \ref{fig:cnn-fool-resnet50-newdef} shows the fooling rates obtained with SVD-Universal with batch size of 64, 128, 1024 on VGG16, VGG19 and ResNet50, respectively. We observed that SVD-Universal attacks obtained using batch size of 64 perform as well as attack vectors obtained using larger batch sizes of 128 and 1024. %We compare our fooling rate results with that of Moosavi-Dezfooli et al~\cite{Dezfooli17} and Khrulkov and Oseledets~\cite{Khrulkov18} on the validation set of ImageNet in Table~\ref{tab:compare-svd-mdfff}\footnote{We would like to point out that the values quoted for Khrulkov-Oseledets in Table~\ref{tab:compare-svd-mdfff} are taken as is from ~\cite[Table 4]{Khrulkov18}}. 
We compare our fooling rate results with that of Moosavi-Dezfooli et al~\cite{Dezfooli17} on the validation set of ImageNet in Table~\ref{tab:compare-svd-mdfff}.

\begin{figure}[!h]
\begin{center}
\includegraphics[width=0.48\linewidth]{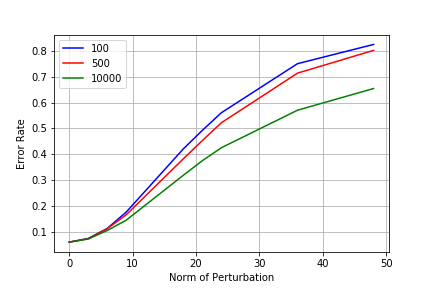}
\includegraphics[width=0.48\linewidth]{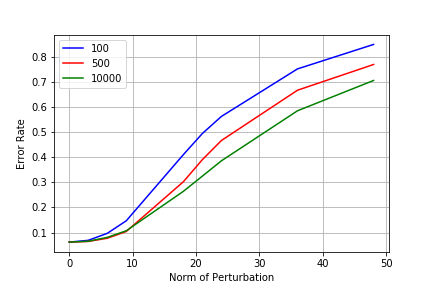}
\includegraphics[width=0.48\linewidth]{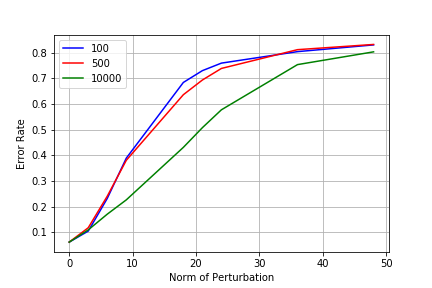}
\end{center}
\vspace{-8pt}
\caption{On CIFAR-10, ResNet18: error rate vs. norm of perturbation along top singular vector of attack directions on 100/500/10000 sample, \textit{(top left)} Gradient \textit{(top right)} FGSM \textit{(bottom)} DeepFool}
\label{fig:cifar10-cnn-fool-fgsm}
\end{figure}

\begin{figure}[!h]
\vspace{-20pt}
\begin{center}
\includegraphics[width=0.55\linewidth]{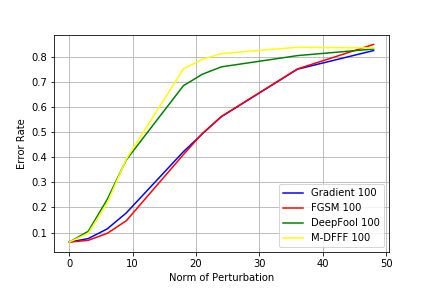}
\end{center}
\vspace{-8pt}
\caption{On CIFAR-10, ResNet18: error rate vs. norm of perturbation along top singular vector of attack directions on 100 samples.}
\label{fig:cifar10-cnn-fool-10}
\end{figure}

%\subsection{SVD-Universal on CIFAR-10}
%\label{app:cifar}
\vspace{6pt}
\noindent \textbf{SVD-Universal on CIFAR-10.}
We plot the error rates of SVD-Universal on CIFAR-10 in Figure \ref{fig:cifar10-cnn-fool-fgsm} trained on ResNet18 with batch size of 100/500/10000. In Figure \ref{fig:cifar10-cnn-fool-10} we plot SVD-Universal and M-DFFF obtained with 100 samples for comparison. Similar to the observation made for ImageNet above, we see that the universal attack with 100 samples performs comparable to the universal attack with larger batch size of 500 and 10000.

%\clearpage

\begin{figure}[h!]
\setlength\tabcolsep{0pt}%%
\centering
\begin{tabular}{cccc}
 \textbf{Norm 0} &
 \textbf{Norm 16} &
 \textbf{Norm 50} &
 \textbf{Norm 100} \\
\includegraphics[width=0.25\linewidth]{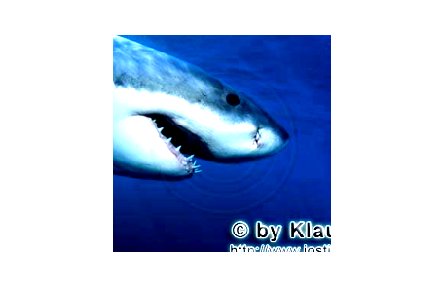} &
\includegraphics[width=0.25\linewidth]{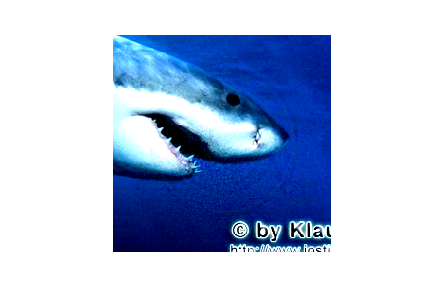} &
\includegraphics[width=0.25\linewidth]{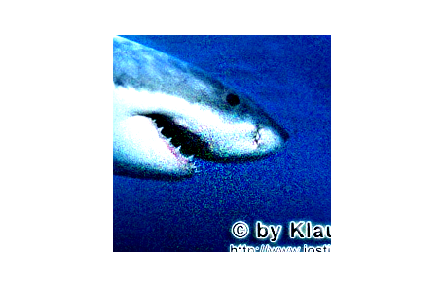} & 
\includegraphics[width=0.25\linewidth]{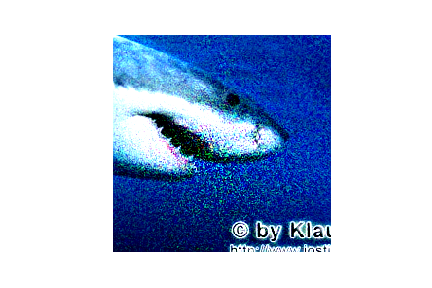} \\
\includegraphics[width=0.25\linewidth]{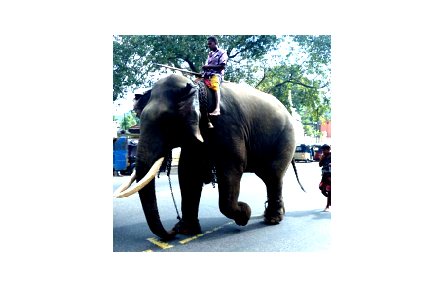} &
\includegraphics[width=0.25\linewidth]{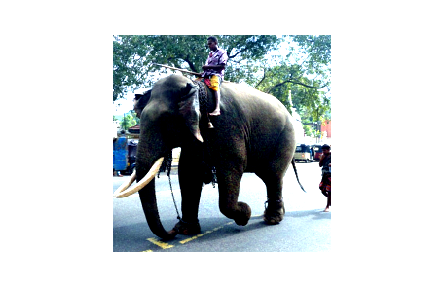} &
\includegraphics[width=0.25\linewidth]{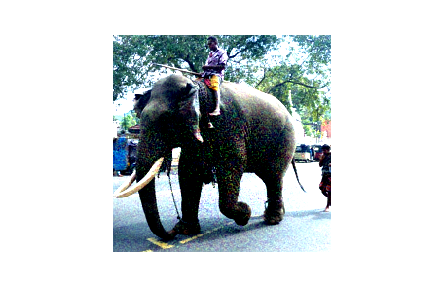} &
\includegraphics[width=0.25\linewidth]{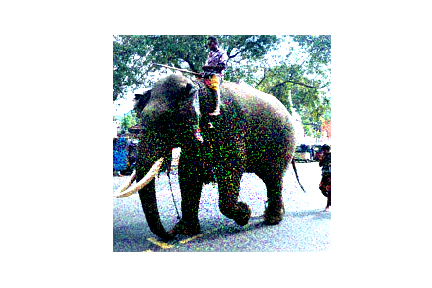} \\
\includegraphics[width=0.25\linewidth]{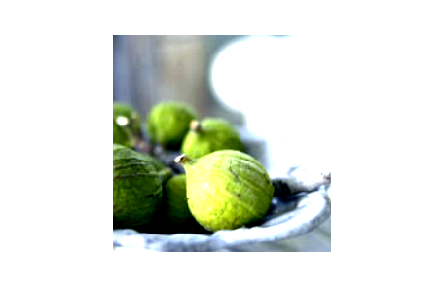} &
\includegraphics[width=0.25\linewidth]{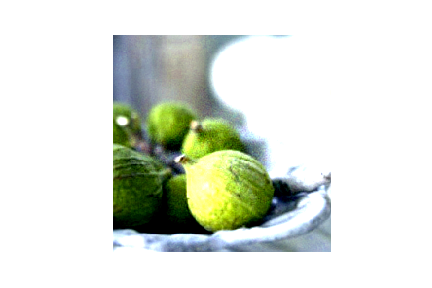} &
\includegraphics[width=0.25\linewidth]{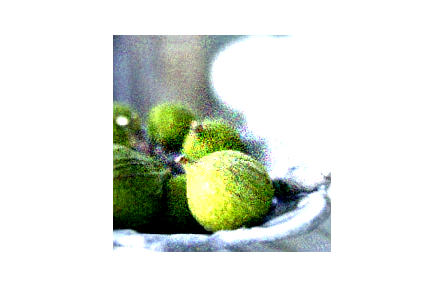} &
\includegraphics[width=0.25\linewidth]{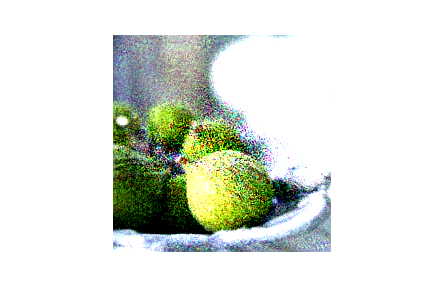} 
\end{tabular}
\caption{Sample images from ImageNet validation set perturbed with SVD-DeepFool of different $l_2$ norms.}
\label{fig:img-103}
\end{figure}

%\eat{Let $v$ denote the attack vector obtained using a batch size of 64. In Figure~\ref{fig:cnn-fool-imagenet-newdef}, we plot the fooling rate of $\epsilon v$ as a function of $\epsilon$, in all the architectures.  The top left plot is the fooling rate of SVD-Gradient, the top right that of SVD-FGSM and the bottom left that of SVD-DeepFool. We explain the bottom right plot in the next section. 
%}
\vspace{6pt}
\noindent \textbf{Our observations.}
\begin{enumerate*}[label=(\roman*)] 
\item We observe a trend similar to what is reported by Khrulkov and Oseledets~\cite{Khrulkov18} - the fooling rate of SVD-Universal attacks is higher on VGG16 and VGG19 than on ResNet50.
\item As noted earlier, we observe that SVD-Universal attacks obtained using batch size of 64 perform as well as attack vectors obtained using larger batch sizes of 128 and 1024.
\item In Khrulkov and Oseledets~\cite[Figure 9]{Khrulkov18}, the authors report that the universal perturbation of ~\cite{Dezfooli17} constructed from a batch size 64 and having $\ell_{\infty}$ norm 10 has a fooling rate of {\bf 0.14} on VGG19. A comparable perturbation in our model has $\ell_2$ norm $4\%$ of 450, and we get a fooling rate of {\bf 0.13} on VGG19.
\item SVD-Gradient attack scaled to have norm 50 has a fooling rate of 0.32 on the validation set of ImageNet for VGG19.   Note that the average $\ell_2$ norm\footnote{For comparison, the average $\ell_2$ norm of the dataset used in ~\cite{Dezfooli17} and ~\cite{Khrulkov18} is 50,000, the average $\ell_{\infty}$ norm is 250, \cite[Footnote, Page 4]{Dezfooli17}. While they use image intensities in the range $[0, 255]$, in our experiments, the pixel intensities are normalized to $[0,1]$, and the average $\ell_2$ norm is 450.} of this validation set\footnote{https://github.com/pytorch/examples/tree/master/imagenet} is 450. 
\item We visualize the perturbed images in Figure~\ref{fig:img-103} when $\epsilon$ is 16 (3.5\% of the average norm of input images) and when $\epsilon$ is 50 (11\% of the average norm of input images). These perturbations are quasi-imperceptible \cite{Dezfooli17}. 
  
\end{enumerate*}

{\small
\begin{table*}[h]
\caption{On ImageNet validation, VGG16 vs VGG19 vs ResNet50 vs M-DFFF: fooling rate vs. norm of perturbation. Attacks constructed using 64 samples. }
\label{tab:compare-svd-mdfff}
\begin{center}
\resizebox{0.65\linewidth}{!} {
\begin{tabular} {| l | l | c | c | c | c |}
\hline
Network  & Vector (using 64 samples) & Norm 18 (4\%)  & Norm 32 (7.1\%)   & Norm 64 (14.2\%) \\
\hline
VGG16    & SVD-Gradient              & 0.12  & 0.19  & 0.34 \\
&     SVD-FGSM                  & 0.10  & 0.17  & 0.31 \\
&     \textbf{SVD-DeepFool}             & \textbf{0.12}  & \textbf{0.20} & \textbf{0.37} \\
&     \textcolor{blue}{M-DFFF}      & \textcolor{blue}{0.11}  & \textcolor{blue}{0.20} & \textcolor{blue}{0.36 }\\
%& \textcolor{magenta}{Khrulkov-Oseledets} &\textcolor{magenta}{0.52} & &\\
\hline
VGG19    & SVD-Gradient              & 0.13  & 0.21  & 0.38 \\
 &   SVD-FGSM                  & 0.10  & 0.17  & 0.30\\
&    \textbf{SVD-DeepFool}              & \textbf{0.11} & \textbf{0.19}  & \textbf{0.33} \\
&    \textcolor{blue}{M-DFFF}                    & \textcolor{blue}{0.11}  & \textcolor{blue}{0.19}  & \textcolor{blue}{0.31} \\
%&   \textcolor{magenta}{Khrulkov-Oseledets} &\textcolor{magenta}{0.60} & &\\
\hline
ResNet50 & SVD-Gradient             & 0.10  & 0.16  & 0.28 \\
&  SVD-FGSM                  & 0.09  & 0.14  & 0.24 \\
&  \textbf{SVD-DeepFool}             & \textbf{0.10}  & \textbf{0.17}  & \textbf{0.29} \\
&  \textcolor{blue}{M-DFFF}                    & \textcolor{blue}{0.09}  & \textcolor{blue}{0.16}  & \textcolor{blue}{0.28} \\
%&  \textcolor{magenta}{Khrulkov-Oseledets} &\textcolor{magenta}{0.44} & &\\
\hline
\end{tabular}
}
\end{center}
\end{table*}
}

%\clearpage

\begin{figure}[]
\centering
\begin{tabular}{ccccc}
 \textbf{VGG16} &
 \textbf{ResNet50} \\
\includegraphics[width=0.44\linewidth]{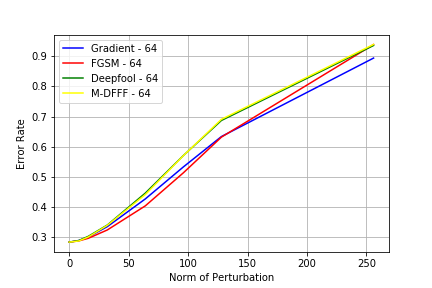}&
\includegraphics[width=0.44\linewidth]{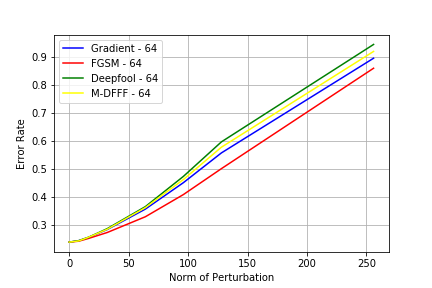}\\
\includegraphics[width=0.44\linewidth]{plots/vgg16_imagenet_0_0_fooling_rate_64_gradvsfgsmvsdp_newdef_notitle.png} &
\includegraphics[width=0.44\linewidth]{plots/resnet50_imagenet_0_0_fooling_rate_64_gradvsfgsmvsdp_newdef_notitle.png}
\end{tabular}
\caption{On ImageNet validation, (top left): VGG16: error rate (bottom left) VGG16: fooling rate (top right) ResNet50: error rate, (bottom right) ResNet50: fooling rate, vs. norm of perturbation along top singular vector of attack directions on 64 samples.}
\label{fig:cnn-fool-vgg16vsres}
\end{figure}

\begin{figure*}[]
\begin{tabular}{cccc}
\textbf{Network} & \textbf{Gradient} & \textbf{FGSM} & \textbf{DeepFool} \\
\begin{sideways}VGG16\end{sideways} 
 & \includegraphics[width=0.24\linewidth]{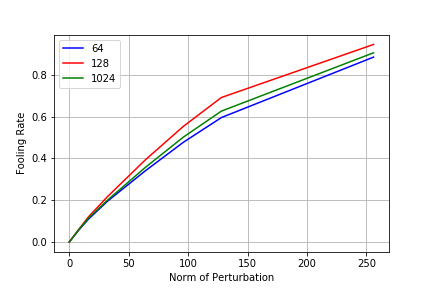}&
   \includegraphics[width=0.24\linewidth]{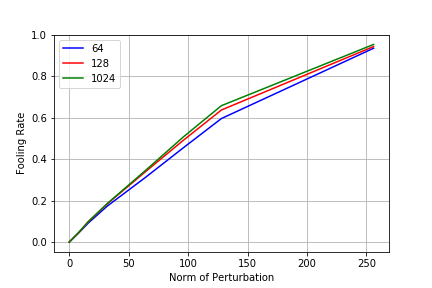}&
   \includegraphics[width=0.24\linewidth]{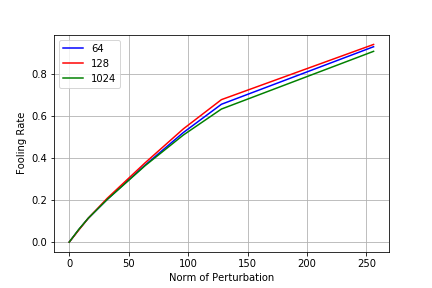}\\
\begin{sideways}VGG19\end{sideways}
 & \includegraphics[width=0.24\linewidth]{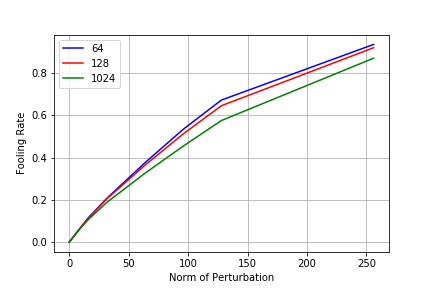}&
   \includegraphics[width=0.24\linewidth]{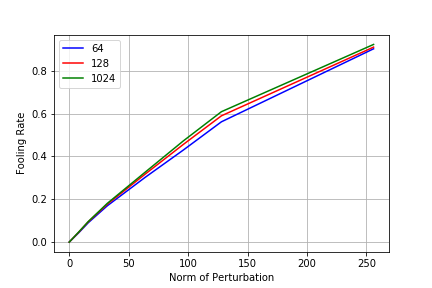}&
   \includegraphics[width=0.24\linewidth]{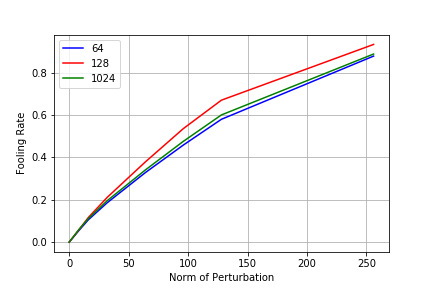}\\
\begin{sideways}ResNet50\end{sideways}
 & \includegraphics[width=0.24\linewidth]{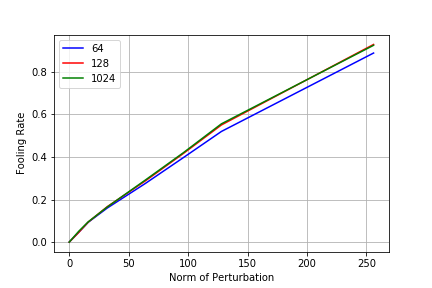}&
   \includegraphics[width=0.24\linewidth]{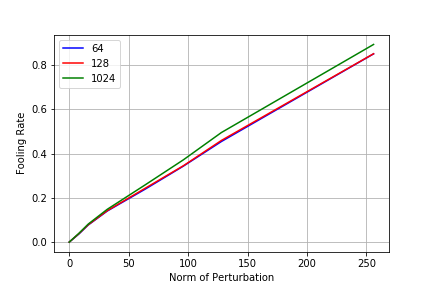}&
   \includegraphics[width=0.24\linewidth]{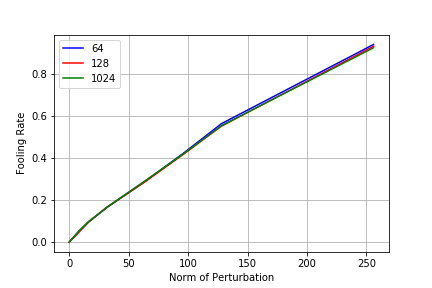}\\
\end{tabular}
\caption{On ImageNet validation, fooling rate as per \cite{Dezfooli17}, on VGG16, VGG19 and ResNet50: fooling rate vs. norm of perturbation along top singular vector of attack directions on 64/128/1024 sample}
\label{fig:cnn-fool-resnet50-newdef}
\end{figure*}
%\clearpage

 %\textcolor{red}{We need to be careful while presenting our comparison with Khrulkov and Oseledets and emphasize that for better fooling rate detailed network analysis will be needed.} 
%The $\ell_2$ norm of the perturbation used  in Moosavi-Dezfooli et al~\cite{Dezfooli17} is $4\%$ of 50,000. The $\ell_{\infty}$ norm of the  universal perturbation used  in Khrulkov and Oseledets ~\cite{Khrulkov18} is $4\%$ of 250 = 10.   \eat{Figure~\ref{fig:cnn-fool-imagenet-newdef}(top right) shows that SVD-Gradient and SVD-DeepFool perturbation with norm 18 give a fooling rate of {\bf 0.13} on VGG19.}
  
 %\eat{In Figure~\ref{fig:cnn-fool-imagenet-newdef} (right bottom) we plot the fooling rate of M-DFFF $\epsilon w$ as a function of $\epsilon$, in all the architectures. } We observe trends similar to what was reported in Moosavi-Dezfooli et al.~\cite[Table 1]{Dezfooli17} - the fooling rate achieved by M-DFFF on VGG16 is higher than that of M-DFFF on VGG19. \eat{Observe from this figure that on VGG19, the M-DFFF universal perturbation of $\ell_2$ norm 18 gives a fooling rate of ${\bf 0.13}$.}

%The plots in Figure~\ref{fig:cnn-fool-imagenet-newdef} of the fooling rates of the four universal attacks on VGG16, VGG19 and ResNet50 are collated together in Figure~\ref{fig:cnn-fool-vgg16vsvgg19vsres}. %We also plot the error rate of these universal attacks on VGG16 and ResNet50.\eat{ in Figure~\ref{fig:cnn-fool-vgg16} (left) and Figure~\ref{fig:cnn-fool-res} (left) .} SVD-DeepFool and M-DFFF have comparable fooling and error rates for ImageNet on the two networks.
 
We note that Khrulkov and Oseledets~\cite{Khrulkov18} get a fooling rate of more than ${\bf 0.4}$ using batch size of 64 and $\ell_{\infty}$ norm 10. Their universal attack is stronger than both our attack and that of Moosavi-Dezfooli et al.~\cite{Dezfooli17}. However, Khrulkov and Oseledets~\cite{Khrulkov18} do an extensive experimentation and determine which intermediate layer to attack and $(p,q)$ are also optimized to maximize the fooling rate of their $(p,q)$-singular vector. $(p,q)$-SVD computation is expensive and is known to be a hard problem, \cite{Bhaskara11,Bhattiprolu19}. We do no such optimization and use $p=q=2$, our emphasis being on the simplicity and universality of our SVD-Universal algorithm.  

\section{Discussion}
%\subsection{Connection between fooling rate and error rate} \label{app:sec:defn}
\noindent \textbf{Connection between fooling rate and error rate.}
As stated earlier, let ${\cal D}$ be a distribution on pairs of images and labels, with images coming from a set ${\mathcal X} \subseteq {\mathbb R}^d$. In the case of CIFAR-10 images, we can think of ${\mathcal X}$ to be the set of $32 \times 32$ CIFAR-10 images with pixel values from $[0, 1]$. So each image is a vector in a space of dimension 1024.
 Let $(X,Y)$ be a sample from ${\cal D}$ and let $f:{\cal X} \rightarrow [k]$, be a $k$-class classifier.
%The natural accuracy $\delta$ of the classifier $f$ is defined to be
%\[ \text{Pr}_{(X, Y) \in {\cal D}} [f(X) = Y]. \]
The error rate of the classifier $f$ is  $\text{Pr}_{(X, Y) \in {\cal D}} [f(X) \not = Y] = \beta$.
 An adversary 
${\cal A}$ is a function ${\cal X} \rightarrow {\mathbb R}^d$. When ${\cal A}$ is a distribution over functions we get a randomized adversary. The norm of the perturbation applied to $X$ is the norm of ${\cal A}(X)$ (we only consider $\ell_2$ norm in this paper).

In Moosavi-Dezfooli et al.~\cite{Dezfooli17,Dezfooli17anal} and Khrulkov and Oseledets~\cite{Khrulkov18}, which we follow in this work for better comparison, the authors consider the fooling rate of an adversary. A classifier $f$ is said to be fooled on input $x$ by the perturbation ${\cal A}(x)$ if $f(x + {\cal A}(x)) \neq f(x)$. The fooling rate of the adversary ${\cal A}$ is defined to be
\[ \text{Pr}_{(X,Y)}[f(X + {\cal A}(X)) \neq f(X)].\]
The adversarial error rate of ${\cal A}$ on the classifier $f$ is defined to be 
$\text{Pr}_{(X, Y) \in {\cal D}}[ f (X + {\cal A}(X)) \neq Y]$. It is easy to see that 
\begin{equation*}
\text{Pr}_{(X,Y)}[ f (X + {\cal A}(X)) \neq f(X)] \geq  \text{Pr}_{(X, Y)}[ f (X + {\cal A}(X)) \neq Y] - \beta. 
%\begin{split}
%& \text{Pr}_{(X,Y)}[ f (X + {\cal A}(X)) \neq f(X)] \\
%& \geq  \text{Pr}_{(X, Y)}[ f (X + {\cal A}(X)) \neq Y] - (1 - \delta). \\
%\end{split}
\end{equation*}
So, if the natural accuracy of the classifier $f$ is high,
the fooling rate is close to the adversarial error rate.
The error rate of the adversary with zero perturbation is the error rate of the trained network, whereas the fooling rate of the adversary with zero perturbation is necessarily zero. However, small fooling rate does not necessarily imply small error rate, especially when the natural accuracy is not close to 100\%. Note that existing models such as VGG16, VGG19, ResNet50 do not achieve natural accuracy greater than 0.8 on the ImageNet dataset. In Figure \ref{fig:cnn-fool-vgg16vsres}, we present the plots of SVD-Universal and M-DFFF on VGG16 and ResNet50 networks using fooling rate and error rate for comparison.

\begin{figure}[!h]
\begin{center}
\includegraphics[width=0.32\linewidth]{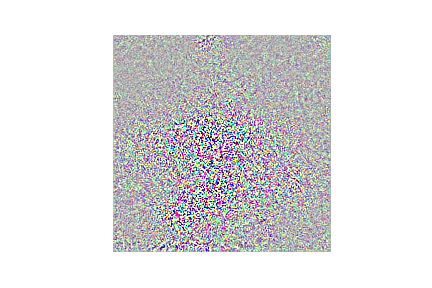}
\includegraphics[width=0.32\linewidth]{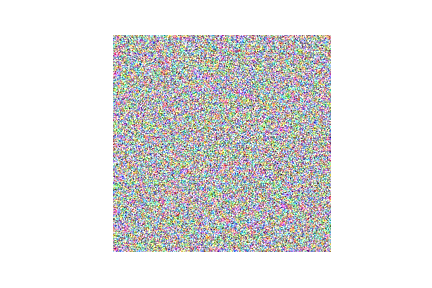}
\includegraphics[width=0.32\linewidth]{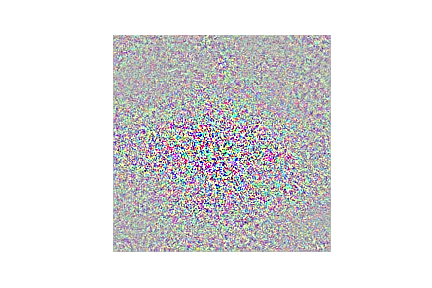}
\end{center}
\caption{For ImageNet, Top SVD vector from (left) Gradient, (center) FGSM, (right) DeepFool on ResNet50.}
\label{imagenet-all-tvh}
\end{figure}

\begin{figure}[!h]
\begin{center}
\includegraphics[width=0.19\linewidth]{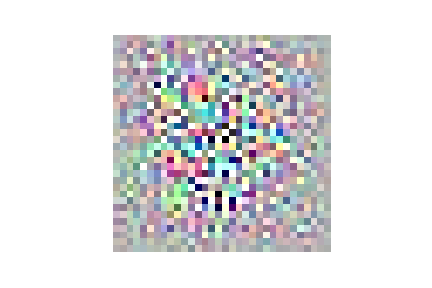}
\includegraphics[width=0.19\linewidth]{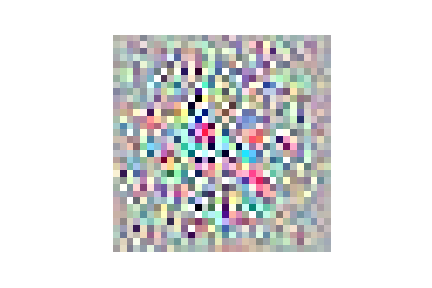}
\includegraphics[width=0.19\linewidth]{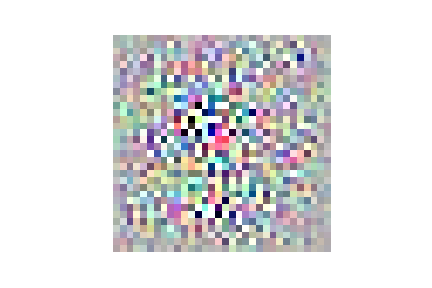}
\includegraphics[width=0.19\linewidth]{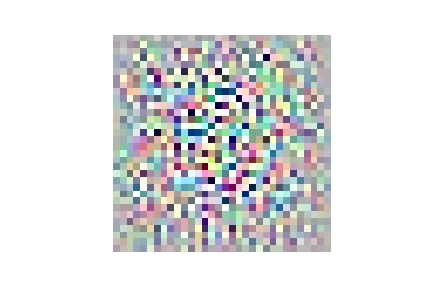}
\includegraphics[width=0.19\linewidth]{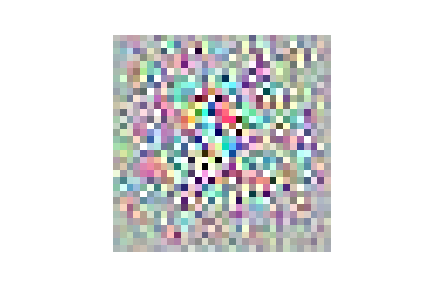}\\
\includegraphics[width=0.19\linewidth]{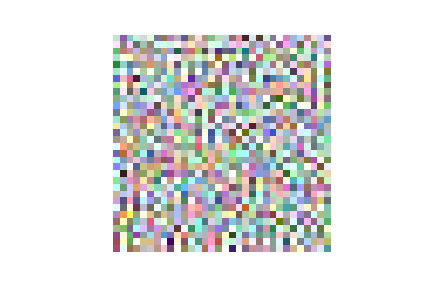}
\includegraphics[width=0.19\linewidth]{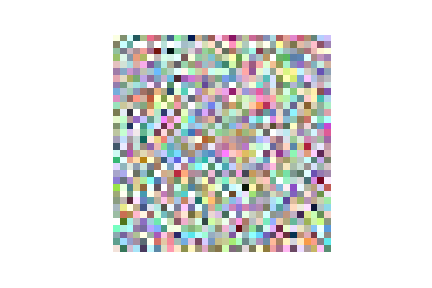}
\includegraphics[width=0.19\linewidth]{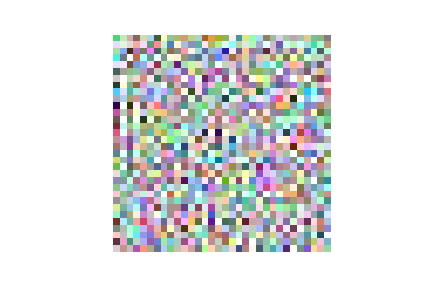}
\includegraphics[width=0.19\linewidth]{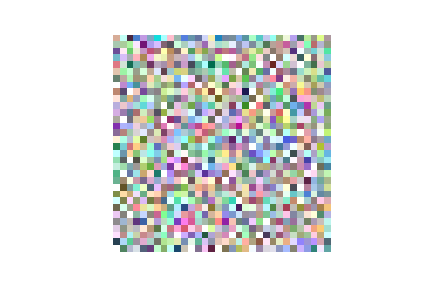}
\includegraphics[width=0.19\linewidth]{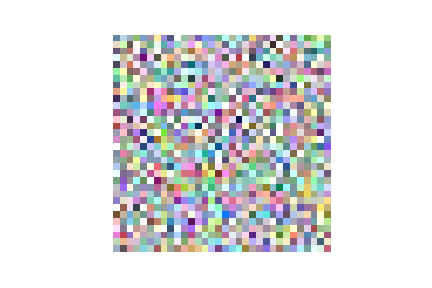}\\
\includegraphics[width=0.19\linewidth]{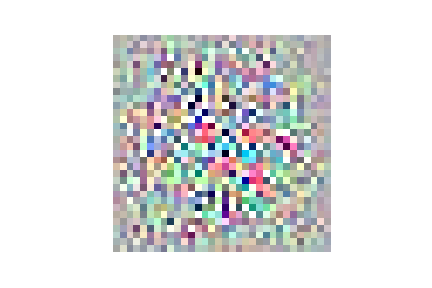}
\includegraphics[width=0.19\linewidth]{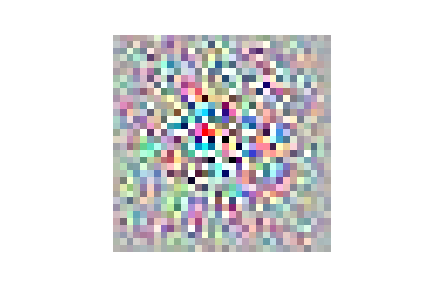}
\includegraphics[width=0.19\linewidth]{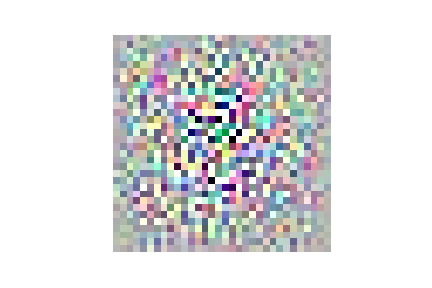}
\includegraphics[width=0.19\linewidth]{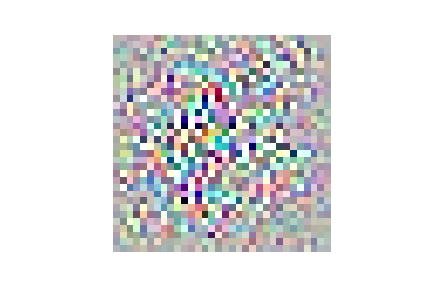}
\includegraphics[width=0.19\linewidth]{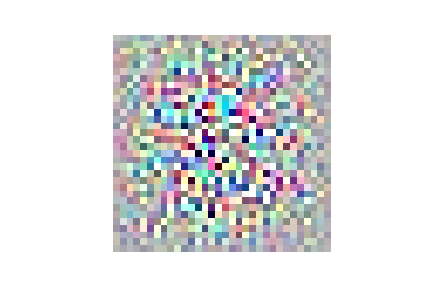}
\end{center}
\caption{For CIFAR-10 (top) Top 5 SVD vectors from {\it Gradient}, (middle) Top 5 SVD vectors from {\it FGSM}, (bottom) Top 5 SVD vectors from DeepFool on ResNet18.}
\label{gcnn-fgsm-tvh}
\end{figure}

%\subsection{Visualizing SVD-Universal perturbations}
%\label{app:visual}
\vspace{6pt}
\noindent \textbf{Visualizing SVD-Universal perturbations.}
%The intensity maps of the top singular vectors for the {\it Gradient}, {\it FGSM}, {\it DeepFool} directions for NNs, StdCNNs have interesting structure. We visualize these perturbations in this section.  These are nearly orthogonal to the invariant directions which lead to steerable filters. We believe this underlying structure is useful and of independent interest. 
We visualize the top singular vectors for the {\it Gradient}, {\it FGSM}, {\it DeepFool} directions from ImageNet and CIFAR-10 in Figure \ref{imagenet-all-tvh} and Figure \ref{gcnn-fgsm-tvh}, respectively. We observe that the {\it Gradient} and {\it DeepFool}-based singular directions are more concentrated in few regions, while the {\it FGSM} is more spread out. This observation could be useful in understanding the role of universal directions and adversarial robustness in general.

%\clearpage

\section{Conclusion}
In this work, we show how to use a small sample of input-dependent adversarial attack directions on test inputs to find a single universal adversarial perturbation that fools state-of-the-art neural network models. Our main observation is a spectral property common to different attack directions such as {\it Gradients}, {\it FGSM}, {\it DeepFool}. We give a theoretical justification for how this spectral property helps in universalizing the adversarial attack directions by using the top singular vector. We justify theoretically and empirically that such a perturbation can be computed using only a small sample of test inputs. 

\noindent \textbf{Acknowledgements.}
Sandesh Kamath would like to thank Microsoft Research India for funding a part of this work through his postdoctoral research fellowship at IIT Hyderabad.

\section{Appendix}

%\begin{appendices}
%\section{Proof details} \label{app:sec:proofs}
The results below are used in the proof of Theorem \ref{thm:sample}, and are included herein for completeness. Theorem 5.6.1 (General covariance estimation) from \cite{vershynin2018} bounds the spectral norm of covariance matrix estimated from a small number of samples as follows.
\begin{theorem}[{\cite[Thm~5.6.1]{vershynin2018}}]
\label{thm:cov-estimation}
Let $X$ be a random vector in $\R^{d}$, $d \geq 2$. Assume that for some $K \geq 1$, $\norm{X} \leq K \left(\expec{\norm{X}^{2}}\right)^{1/2}$ , almost surely. Let $\Sigma = \expec{XX^{T}}$ be the covariance matrix of $X$ and $\Sigma_{m} = \frac{1}{m} \sum_{i=1}^{m} X_{i} X_{i}^{T}$ be the estimated covariance from $m$ i.i.d. samples $X_{1}, X_{2}, \dotsc, X_{m}$. Then for every positive integer $m$, we have 
\[
\expec{\norm{\Sigma_{m} - \Sigma}} \leq C \left(\sqrt{\frac{K^{2} d \log d}{m}} + \frac{K^{2} d \log d}{m}\right) \norm{\Sigma},
\]
for some positive constant $C$ and $\norm{\Sigma}$ being the spectral norm (or the top eigenvalue) of $\Sigma$. 
\end{theorem}
Note that using $m = O(\epsilon^{-2} d \log d)$ we get $\expec{\norm{\Sigma_{m} - \Sigma}} \leq \epsilon \norm{\Sigma}$.
A tighter version of Theorem 5.6.1 appears as Remark 5.6.3, when the \emph{intrinsic dimension} $r = \tr{\Sigma}/\norm{\Sigma} \ll d$.
\begin{theorem}[{\cite[Remark~5.6.3]{vershynin2018}}] 
\label{thm:cov-estimation-intrinsic}
Let $X$ be a random vector in $\R^{d}$, and $d \geq 2$. Assume that for some $K \geq 1$, $\norm{X} \leq K \left(\expec{\norm{X}^{2}}\right)^{1/2}$ , almost surely. Let $\Sigma = \expec{XX^{T}}$ be the covariance matrix of $X$ and $\Sigma_{m} = \frac{1}{m} \sum_{i=1}^{m} X_{i} X_{i}^{T}$ be the estimated covariance from $m$ i.i.d. samples $X_{1}, X_{2}, \dotsc, X_{m}$. Then for every positive integer $m$, we have 
\[
\expec{\norm{\Sigma_{m} - \Sigma}} \leq C \left(\sqrt{\frac{K^{2} r \log d}{m}} + \frac{K^{2} r \log d}{m}\right) \norm{\Sigma},
\]
for some positive constant $C$ and $\norm{\Sigma}$ being the operator norm (or the top eigenvalue) of $\Sigma$. 
\end{theorem}
Note that using $m = O(\epsilon^{-2} r \log d)$ we get $\expec{\norm{\Sigma_{m} - \Sigma}} \leq \epsilon \norm{\Sigma}$.
Theorem 4.5.3 (Weyl's Inequality) from \cite{vershynin2018} upper bounds the difference between $i$-th eigenvalues of two symmetric matrices $A$ and $B$ using the spectral norm of $A - B$.
\begin{theorem}[{\cite[Thm~4.5.3 (Weyl's Inequality)]{vershynin2018}}] 
\label{thm:weyl}
For any two symmetric matrices $A$ and $B$ in $\R^{d \times d}$, $\abs{\lambda_{i}(A) - \lambda_{i}(B)} \leq \norm{A - B}$, where $\lambda_{i}(A)$ and $\lambda_{i}(B)$ are the $i$-th eigenvalues of $A$ and $B$, respectively.
\end{theorem}
In other words, the spectral norm of matrix perturbation bounds the stability of its spectrum.
Here is a special case of Theorem 4.5.5 (Davis-Kahan Theorem) and its immediate corollary mentioned in \cite{vershynin2018}.
\begin{theorem}[{\cite[Thm~4.5.5 (Davis-Kahan Theorem)]{vershynin2018}}] 
\label{thm:davis-kahan}
Let $A$ and $B$ be symmetric matrices in $\R^{d \times d}$. Fix $i \in [d]$ and assume that the largest eigenvalue of $A$ is well-separated from the rest of the spectrum, that is, $\lambda_{1}(A) - \lambda_{2}(A) \geq \delta > 0$. Then the angle $\theta$ between the top eigenvectors $v_{1}(A)$ and $v_{1}(B)$ of $A$ and $B$, respectively, satisfies $\sin \theta \leq 2 \norm{A - B}/\delta$.
\end{theorem}
As an easy corollary, it implies that the top eigenvectors $v_{1}(A)$ and $v_{1}(B)$ are close to each other up to a sign, namely, there exists $s \in \{-1, 1\}$ such that
\[
\norm{v_{1}(A) - s~ v_{1}(B)} \leq \frac{2^{3/2} \norm{A - B}}{\delta}.
\]

%%
%% The next two lines define the bibliography style to be used, and
%% the bibliography file.
%\clearpage
\bibliographystyle{splncs04}
\bibliography{universal}

\begin{thebibliography}{10}
\providecommand{\url}[1]{\texttt{#1}}
\providecommand{\urlprefix}{URL }
\providecommand{\doi}[1]{https://doi.org/#1}

\bibitem{Bhaskara11}
Bhaskara, A., Vijayaraghavan, A.: Approximating matrix p-norms. In: Proceedings
  of the Twenty-second Annual ACM-SIAM Symposium on Discrete Algorithms. pp.
  497--511. SODA '11, Society for Industrial and Applied Mathematics,
  Philadelphia, PA, USA (2011)

\bibitem{Bhattiprolu19}
Bhattiprolu, V., Ghosh, M., Guruswami, V., Lee, E., Tulsiani, M.:
  Approximability of p {$\rightarrow$} q matrix norms: Generalized krivine
  rounding and hypercontractive hardness. In: Proceedings of the Thirtieth
  Annual ACM-SIAM Symposium on Discrete Algorithms. pp. 1358--1368. SODA '19,
  Society for Industrial and Applied Mathematics, Philadelphia, PA, USA (2019)

\bibitem{Goodfellow15}
Goodfellow, I.J., Shlens, J., Szegedy, C.: Explaining and harnessing
  adversarial examples. In International Conference on Learning Representations
   (2015)

\bibitem{He16}
He, K., Zhang, X., Ren, S., Sun, J.: Deep residual learning for image
  recognition. In Proceedings of the IEEE conference on computer vision and
  pattern recognition pp. 770--778 (2016)

\bibitem{Khrulkov18}
Khrulkov, V., Oseledets, I.: Art of singular vectors and universal adversarial
  perturbations. In: The IEEE Conference on Computer Vision and Pattern
  Recognition (CVPR) (June 2018)

\bibitem{Kurakin17}
Kurakin, A., Goodfellow, I., Bengio, S.: Adversarial examples in the physical
  world. arXiv preprint arXiv:1607.02533  (2017)

\bibitem{Madry18}
Madry, A., Makelov, A.A., Schmidt, L., Tsipras, D., Vladu, A.: Towards deep
  learning models resistant to adversarial attacks. In International Conference
  on Learning Representations  (2018)

\bibitem{Dezfooli17}
Moosavi-Dezfooli, S.M., Fawzi, A., Fawzi, O., Frossard, P.: Universal
  adversarial perturbations. In Proceedings of the IEEE Conference on Computer
  Vision and Pattern Recognition  (2017)

\bibitem{Dezfooli17anal}
Moosavi{-}Dezfooli, S., Fawzi, A., Fawzi, O., Frossard, P., Soatto, S.:
  Analysis of universal adversarial perturbations. arXiv preprint
  arXiv:1705.09554  (2017)

\bibitem{Dezfooli16}
Moosavi-Dezfooli, S.M., Fawzi, A., Frossard, P.: Deepfool: a simple and
  accurate method to fool deep neural networks. In Proceedings of the IEEE
  Conference on Computer Vision and Pattern Recognition  (2016)

\bibitem{Mopuri17}
Mopuri, K.R., Garg, U., Babu, R.V.: Fast feature fool: A data independent
  approach to universal adversarial perturbations. In: Proceedings of the
  British Machine Vision Conference ({BMVC}) (2017)

\bibitem{Russakovsky2015}
Russakovsky, O., Deng, J., Su, H., Krause, J., Satheesh, S., Ma, S., Huang, Z.,
  Karpathy, A., Khosla, A., Bernstein, M., Berg, A.C., Fei-Fei, L.: Imagenet
  large scale visual recognition challenge. International Journal of Computer
  Vision  \textbf{115}(3),  211--252 (Dec 2015)

\bibitem{Szegedy13}
Szegedy, C., Zaremba, W., Sutskever, I., Bruna, J., Erhan, D., Goodfellow,
  I.J., Fergus, R.: Intriguing properties of neural networks. arXiv preprint
  arXiv:1312.6199  (2013)

\bibitem{Tramer17}
Tramer, F., Papernot, N., Goodfellow, I., Boneh, D., McDaniel, P.: The space of
  transferable adversarial examples. arXiv preprint arXiv:1704.03453  (2017)

\bibitem{vershynin2018}
Vershynin, R.: High-Dimensional Probability: An Introduction with Applications
  in Data Science. Cambridge Series in Statistical and Probabilistic
  Mathematics, Cambridge University Press (2018)

\end{thebibliography}

%%
%% If your work has an appendix, this is the place to put it.
%\appendix
%\end{appendices}
 
\end{document}